# A Non-Parametric Control Chart for High Frequency Multivariate Data

Deovrat Kakde dev.kakde@sas.com, Sergiy Peredriy sergiy.peredriy@sas.com, Arin Chaudhuri arin.chaudhuri@sas.com, Anya McGuirk anya.mcguirk@sas.com

## SUMMARY & CONCLUSIONS

Support Vector Data Description (SVDD) is a machine learning technique used for single class classification and outlier detection. A SVDD based *K*-chart was first introduced by Sun and Tsung for monitoring multivariate processes when the underlying distribution of process parameters or quality characteristics departed from normality. The method first trains a SVDD model on data obtained from stable or in-control operations of the process to obtain a threshold $R^2$ and kernel center *a*. For each new observation, its Kernel distance from the Kernel center *a* is calculated. The kernel distance is compared against the threshold $R^2$ to determine if the observation is within the control limits.

The non-parametric *K*-chart provides an attractive alternative to the traditional control charts such as the Hotelling's $T^2$ charts when the distribution of the underlying multivariate data is either non-normal or is unknown. But there are challenges when the *K*-chart is deployed in practice. The *K*-chart requires calculating the kernel distance of each new observation but there are no guidelines on how to interpret the kernel distance plot and draw inferences about shifts in process mean or changes in process variation. This limits the application of *K*-charts in big-data applications such as equipment health monitoring, where observations are generated at a very high frequency. In this scenario, the analyst using the *K*-chart is inundated with kernel distance results at a very high frequency, generally without any recourse for detecting presence of any assignable causes of variation.

We propose a new SVDD based control chart, called a $K_T$ chart, which addresses the challenges encountered when using a *K*-chart for big-data applications. The $K_T$ charts can be used to track simultaneously process variation and central tendency.

We illustrate the successful use of $K_T$ chart using the Tennessee Eastman process data.

## 1 INTRODUCTION

Real-time monitoring of equipment health for proactive intervention is one of the most promising areas within the broader realm of Internet of Things (IoT). Such measurement data is multivariate and often generated at a very high frequency. For example, a typical airplane currently has ~ 6,000 sensors measuring critical health parameters and creates 2.5 terabytes of data per day [1]. Additionally, since such equipment are inherently reliable, the majority of measurements correspond to a healthy or normal state. Multivariate Statistical Process Control charts are used to monitor such data. In phase I, the historical measurement data corresponding to the normal state is used to define the state of statistical control. The new measurement data is monitored in phase II for departure from the state of statistical control and possible presence of assignable causes.

The first parametric control chart for multivariate data was introduced by Hotelling in 1947 [2]. It introduced the use of the $T^2$ statistic for monitoring multivariate data. Kourti and MacGregor introduced principle component based multivariate control charts for applications involving large number of correlated variables [3]. Both charts assume that the process variables follow a multivariate normal distribution. Sun and Tsung proposed the SVDD based *K*-charts for monitoring multivariate processes when underlying distribution of process or health parameters depart from normality [4]. The use of SVDD allows the control limit to be non-ellipsoid and conform to the actual shape of the data. In this paper, we outline the shortcomings of this *K*-chart for monitoring high frequency multivariate data. We propose a SVDD based $K_T$ chart, which addresses the shortcomings of the *K*-chart.

## 2 SUPPORT VECTOR DATA DESCRIPTION

SVDD is a machine learning technique used for single class classification and outlier detection. The SVDD technique is similar to Support Vector Machines and was first introduced by Tax and Duin [5]. SVDD is used in domains where the majority of data belongs to a single class.

2.1 *Mathematical Formulation of SVDD (see [5])*

*Normal Data Description:*

The SVDD model for normal data description builds a minimum radius hypersphere around the data.

*Primal Form:*

Objective Function:

$$\min R^2 + C \sum_{i=1}^{n} \xi_i \quad (1)$$

Subject to:
$$\|x_i - a\|^2 \leq R^2 + \xi_i, \forall i = 1, \ldots, n \quad (2)$$
$$\xi_i \geq 0, \forall i = 1, \ldots, n \quad (3)$$

Where:
$x_i \in \mathbb{R}^m, i = 1, \ldots, n$ represents the training data
$R$: radius represents the decision variable
$\xi_i$: is the slack of each variable,
$a$: is the center, a decision variable
$C = \frac{1}{nf}$: is the penalty constant that controls the trade-off between the volume and the errors, and;
$f$: is the expected outlier fraction

*Dual Form:*

The dual formulation is obtained using the Lagrange multipliers.
Objective Function:
$$\max \sum_{i=1}^{n} \alpha_i (x_i . x_j) - \sum_{i,j} \alpha_i \alpha_j (x_i . x_j) \quad (4)$$
Subject to:
$$\sum_{i=1}^{n} \alpha_i = 1 \quad (5)$$
$$0 \leq \alpha_i \leq C, \forall i = 1, \ldots, n \quad (6)$$

Where:
$\alpha_i \in \mathbb{R}, i = 1, \ldots, n$ are the Lagrange coefficients

*Duality Information:*

Depending upon the position of the observation, the following results hold good:
Position of the Center $a$: $\sum_{i=1}^{n} \alpha_i x_i$ (7)
Inside Position: $\|x_i - a\| < R \rightarrow \alpha_i = 0$ (8)
Boundary Position: $\|x_i - a\| = R \rightarrow 0 < \alpha_i < C$ (9)
Outside Position: $\|x_i - a\| > R \rightarrow \alpha_i = C$ (10)

The circular data boundary can include a significant amount of space with a very sparse distribution of training observations. Scoring with this model can increase the probability of false positives. Hence, instead of a circular shape, a compact bounded outline around the data is often used. Such an outline should approximate the shape of the single-class training data. This is possible with the use of kernel functions.

The SVDD is made flexible by replacing the inner product $(x_i . x_j)$ with a suitable kernel function $K(x_i, x_j)$. The Gaussian kernel function used in this paper is defined as:
$$K(x_i, x_j) = \exp \frac{-\|x_i - x_j\|^2}{2s^2} \quad (11)$$

Results 7 through 10 hold good when the kernel function is used in the mathematical formulation. Using the Kernel function, the threshold $R^2$ is calculated as:
$$R^2 = K(x_k, x_k) - 2 \sum_i (x_i, x_k) - \sum_{i,j} \alpha_i \alpha_j (x_i, x_j) \quad (12)$$
using any $x_k \in SV_{<C}$ where $SV_{<C}$ is the set of support vectors that have $\alpha_k < C$.

*Scoring:*
For each observation $z$ in the scoring dataset, the distance $dist^2(z)$ is calculated as follows:
$$dist^2(z) = K(z, z) - 2 \sum_i (x_i, z) - \sum_{i,j} \alpha_i \alpha_j (x_i, x_j) \quad (13)$$

The scoring dataset points with $dist^2(z) > R^2$ are designated as outliers.

### 2.2 SVDD-based Control Charts

Sun and Tsung first introduced SVDD based $K$-charts for monitoring multivariate observations. In Phase I, the $K$-chart uses a SVDD training algorithm and develops a model of a process under statistical control. The Phase I statistics obtained using the 'in-control' data include the set of support vectors $SV$, the set of corresponding Lagrange coefficients $\alpha$, the threshold $R^2$ value and the center $a$. The Phase II control limits are set as $LCL=0$ and $UCL=R^2$. The sets $SV$ and $\alpha$ are used in Phase II to compute distance $dist^2(z)$, which represents the distance of a new observation $z$ from the center $a$. The value of $dist^2(z)$ is compared to the $UCL$. The observations with $dist^2(z)$ greater than the $UCL$ are outliers and are examined for presence of any assignable cause(s).

$K$-chart provides a good alternative to the traditional $T^2$ based control charts when the underlying distribution is either non-normal or unknown. $K$-chart is an individual value control chart thus it cannot be readily used to identify a shift in the process center or a change in the process spread. This potentially limits applications of $K$- charts in IoT domains, where data is multivariate and generated at a very high frequency. In such applications, monitoring $dist^2(z)$ value of each observation and, based on that, drawing any meaningful inference about the process behavior would involve additional computations and can be a challenge, especially when data is generated at a very high frequency.

### 3 $K_T$ CHARTS

We propose SVDD based $K_T$ charts for monitoring high frequency multivariate data. The $K_T$ charts use sliding windows to define observation subgroups. For each window, we develop an SVDD model and obtain corresponding center $a$ and threshold $R^2$. We use center $a$ as a measure of the window's central tendency and threshold $R^2$ as a measure of the window's variation. The $K_T$ chart allows simultaneous monitoring of both, the central tendency and the spread of the process. The $K_T$ chart for monitoring process center is called the $a$ chart, whereas the $K_T$ chart for monitoring variation is called the $R^2$ chart.

The SVDD training involves solving a quadratic programming problem and the time required to obtain a solution is directly related to the number of observations in the time window. Hence, SVDD training is computationally expensive when data is generated at a high frequency and/or the number of observations in the window is very high. We solve this issue by using a sampling based method for fast SVDD training. The method develops a SVDD model for each window by computing SVDD using independent random samples selected with replacement from the window. The method is extremely fast, robust to outliers and provides a near-identical data description compared with training the entire data set in one iteration.

In the remainder of this section, we first outline design of $K_T$, which is then followed by an explanation of the sampling-based method and the guidelines for selecting the parameters of the $K_T$ charts.

### 3.1 Design of $K_T$ Charts

*Notations:*
$T$: phase I data
$p$: number of observations in data set $T$
$j$: index of observations in data set $T$, $j = 1$ to $p$.
$n$: window length used in phase I and II.
$m$: window overlap, where $0 \leq m \leq n - 1$
$W_i$: $i=1, 2... k$ represents the $i^{th}$ window, where $k = \left\lfloor \frac{p}{n-m} \right\rfloor$
$START(W_i)$: First observation of window $W_i$.
$START(W_i) = \begin{cases} 1 & \text{if } i = 1, \\ (i-1) \times n - m + 1 & \text{otherwise} \end{cases}$
$END(W_i)$: last observation of window $W_i$.
$END(W_i) = START(W_i) + n$
$\langle SV_i, R_i^2, a_i \rangle \leftarrow TRAIN_{SAMP}(W_i)$: indicates the set of support vectors $SV_i$, the threshold value $R_i^2$ and the center $a_i$ obtained by performing SVDD computations using the sampling based method on window $W_i$.
$A$: data set containing the center $a_i$ of each window, where $i = 1$ to $k$
$\langle SV_a, \alpha_a, R_a^2, a^* \rangle \leftarrow TRAIN(A)$: indicate the set of support vectors $SV_a$, set of corresponding Lagrange coefficients $\alpha_a$, threshold value $R_a^2$ and center $a^*$ obtained by performing SVDD computations on data set $A$ using all observations in $A$ in one iteration.
$SCORE(a_q; SV_a, \alpha_a)$: indicate scoring a center $a_q$ in phase II, using $\langle SV_a, \alpha_a R_a^2 \rangle$
$dist^2(a_q) \leftarrow SCORE(a_q; SV_a, \alpha_a)$: indicates distance of center $a_q$ corresponding to the window $W_q$ from center $a^*$ obtained by scoring center $a_q$ using $\langle SV_a, \alpha_a \rangle$
Note that $a_q$ represents a center of window $w_q$ obtained using $TRAIN_{SAMP}(W_q)$

### $K_T$ Charts: Phase I

The steps for constructing $K_T$ charts in Phase I are described below.
*Step 1:* Obtain description of each window in Phase I data using the sampling based method.
$\langle SV_i, R_i^2, a_i \rangle \leftarrow TRAIN_{SAMP}(W_i)$; for $i=1$ to $k$
*Step 2:* Obtain $\overline{R^2}$ and $\sigma_{R^2}$
$\overline{R^2} = \frac{\sum_{i=1}^{k} R_i^2}{k}$, $\sigma_{R^2} = \sqrt{\frac{\sum_{i=1}^{k} R_i^2 - \overline{R^2}}{k-1}}$
*Step 3:* Perform SVDD computations on data set $A$.
$\langle SV_a, \alpha_a R_a^2, a^* \rangle \leftarrow TRAIN(A)$
*Step 4:* Construct control limits for the $a$ chart and the $R^2$ chart.

> *Control Limits for the a chart*
> $UCL = R_a^2$
> Center Line = $\frac{R_a^2}{2}$
> $LCL = 0$

> *Control Limits for the $R^2$ chart*
> $UCL = \overline{R^2} + 3 \times \sigma_{R^2}$
> Center Line = $\overline{R^2}$
> $LCL = \overline{R^2} - 3 \times \sigma_{R^2}$

*Note:* Optionally, Upper Warning Limits (UWL) and Lower Warning Limits (LWL) for the $R^2$ chart can be set at $\overline{R^2} + 2 \times \sigma_{R^2}$ and $\overline{R^2} - 2 \times \sigma_{R^2}$ respectively.
*Step 5:* Obtain $dist^2(a_i)$, the distance of $a_i$ from $a^*$ as
$dist^2(a_i) \leftarrow SCORE(a_i; SV_a, \alpha_a); \forall i$
*Step 6:* Plot $dist^2(a_i)$ on the $a$ chart and plot $R_i^2$ on the $R^2$ chart $\forall i$
*Step 7:* Investigate the $a$ and $R^2$ charts for presence of potential assignable causes. Drop points corresponding to confirmed presence of assignable causes. Repeat steps 2 through 6, until a state of statistical control is obtained.

### $K_T$ Charts: Phase II

The steps for constructing $K_T$ charts in Phase II are described below.
*Step 1:* Determine $START(W_i)$, the starting point of the $q^{th}$ window, where $q \geq 1$. Obtain window $W_q$ by accumulating $n$ contiguous observations starting from $START(W_q)$.
*Step 2:* Perform SVDD computations on window $W_q$ using the sampling based method.
$\langle SV_q, R_q^2, a_q \rangle \leftarrow TRAIN_{SAMP}(W_q)$; for q $\geq 1$
Plot $R_q^2$ on the $R^2$ chart.
*Step 3:* Compute distance of center $a_q$ from $a^*$ using $SV_a$ and $\alpha_a$
$dist^2(a_q) \leftarrow SCORE(a_q; SV_a, \alpha_a)$
Plot $dist^2(a_q)$ on the $a$ chart.
*Step 4:* Repeat steps 1 through 3 as the new window is populated.

### 3.2 Sampling-based Method for SVDD training:

As outlined in Section 3, the sampling-based method for SVDD training is at the heart of the $K_T$-charts. The sampling-based method provides an efficient way to train SVDD model [6]. Fast training allows us to use observation windows and characterize each window using center $a$ and threshold $R^2$. We observed magnitude performance gains using the sampling method as compared to training using all window observations in one iteration.

The method iteratively samples from the training data set with the objective of updating a set of support vectors called the master set of support vectors, $SV^*$. During each iteration, the method updates $SV^*$ and the corresponding threshold $R^2$ value and center $a$. As the threshold value $R^2$ increases, the volume enclosed by the $SV^*$ increases. The method stops iterating and provides a solution when the threshold value $R^2$, center $a$ and hence the volume enclosed by $SV^*$, converges. At convergence, the members of the master set of support vectors $SV^*$ characterize the description of the training data set. The sampling-based method provides a good approximation to the solution that can be obtained by using all observations in the training data set. Table 1 summarizes the sampling-based method.

---

Requires: Training data set, sample size $n$, Gaussian bandwidth parameter s and fraction outlier $f$

---

*Step 1:* Initialize a master set of support vectors $SV^*$, by computing SVDD of random sample $S_0$. Set $i=1$.

*Step 2:* Select a random sample $S_i$. Compute its SVDD. Obtain a set of support vectors $SV_i$.

*Step 3:* Compute SVDD of $(SV^* \cup SV_i)$. Designate corresponding support vectors as $SV^*$. Set $i=i+1$.

*Step 4:* Repeat step 2 and 3 till convergence of threshold value $R^2$ and center $a$.

---

*Table 1-Sampling –based method for SVDD Training*

### 3.3 Parameter Selection Guidelines:

This section provides some guidelines for setting appropriate values for the $K_T$-chart parameters.

*Window length n and overlap m:* Increasing window length decreases the number of points plotted on the $K_T$ chart and vice versa. When observations are generated at a high frequency and the process is predominantly stable, large window sizes can be used. For unstable processes, with lots of fluctuations, the window size needs to be smaller. Increasing window overlap increases the probability of early discovery of changes in the process center and variation. For a given window length, increasing the window overlap increases the number of windows. Window length and overlap should be selected in concert by taking into account process stability, fault initiation process, expected fault duration and the extent of early warning desired.

*Gaussian bandwidth parameter s:* In the case of a Gaussian kernel, it is observed that if the value of outlier fraction $f$ is kept constant, the number of support vectors identified by the SVDD algorithm is a function of the Gaussian bandwidth parameter $s$. At a very low value of $s$, the number of support vectors is very high, approaching the number of observations. As the value of $s$ increases, the number of support vectors reduces. It is also observed that at lower values of $s$, the data boundary is extremely wiggly.

As $s$ is increased, the data boundary becomes less wiggly, and it starts to follow shape of the data. At higher values of $s$, the data boundary starts becoming spherical.

In support vector machines, cross-validation is a widely used technique for selecting the Gaussian bandwidth parameter. Cross-validation requires training data that belongs to multiple classes. A typical application of $K_T$-charts involves monitoring sensor data obtained from equipment or processes, which are reliable or stable for the most part. Generally, a good amount of data related to normal operations is available, but the data coming out of unstable operations or faulty conditions is either unavailable or not labeled. Hence, cross-validation is not a feasible technique for selecting Gaussian bandwidth parameter value in $K_T$-charts.

The unsupervised methods for selecting Kernel bandwidth parameter do not require labelled data. These methods are suitable for $K_T$-charts. A review of these methods is provided in [7].

*Fraction Outlier $f$:* The full SVDD method is sensitive to the value of the fraction outlier $f$. The value of $f$ is inversely related to the penalty constant $C$ used to control the trade-off between the volume enclosed by the hypersphere and the error, as outlined in the primal formulation of SVDD. Equations 9 and 10 indicate that higher $f$ value, which translates to lower the $C$ value, constrains the $\alpha_i$ more; hence more observations are designated as support vectors. Lower C values decrease the volume enclosed by the SVDD boundary. A data description obtained using an incorrect value of $f$ can be misleading. If the specified value of $f$ is more than its actual value, it can lead to a higher misclassification rate, as observations are incorrectly classified as outliers. The corresponding threshold $R^2$ value will be smaller as well. On the other hand, if the specified value of $f$ is less than its actual value, it can lead to a higher misclassification rate as well, by classifying outliers as inliers.

The sampling based method is robust to the error in estimating $f$. The sample size used in the sampling based method is generally a very small fraction of the training data set. Hence the corresponding value of $C$ used in each iteration is sufficiently large. The strategy of random sampling with replacement with a small sample size lowers the probability of a sample containing any outlier observations. A higher $C$ value ensures that, as the sampling method iteratively expands the data description, the sampled points are not incorrectly classified as outliers.

For $K_{T-}$ chart construction, it is recommended to use domain expertise and select phase I data such that it truly reflects the normal operations. The corresponding $f$ value can be set to a very small value.

*Sample Size:* The performance of sampling method is evaluated using various sample sizes in [7]. For each input dataset, there is an optimal value of sample size which provides lowest total processing time. Although there is no significant difference in performance across different sample sizes, as mentioned in [7], we recommend selecting sample size as $v+1$, where $v$ is the number of variables. This sample size provides a fairly good performance with respect to the processing time.

## 4 MONITORING TENNESSEE EASTMAN PROCESS USING $K_T$ CHARTS

The Tennessee Eastman Process (TE) is a realistic model of a chemical process widely used in the academic community for studying multivariate process control problems [8]. The TE process is based on an actual industrial process, with some modifications to protect the proprietary nature of the process. We used MATLAB simulation code provided by [9] to generate the TE process data. We generated data for the normal operations of the process and twenty different faulty conditions. Each observation consists of 41 variables, out of which 22 are measured continuously, on average, every 6 seconds and the remaining 19 sampled at a specified interval either every 0.1 or 0.25 hours. We interpolated the 22 observations which are measured continuously using the SAS® EXPAND procedure. The interpolation increased the observation frequency, generated 20 observations per second and ensured that we have an adequate data volume to evaluate $K_T$ charts.

The Phase I data contained 125,352 observations belonging to the normal operations of the process. We interleaved the fault data with the normal operations data, as illustrated in Figure 1, to create the Phase II data. The Phase II data contained 4,598,620 observations. The window size was set to 10,000 with an overlap of 3,000. The Gaussian bandwidth parameter was computed as 25.5 using the Peak criteria outlined in [7]

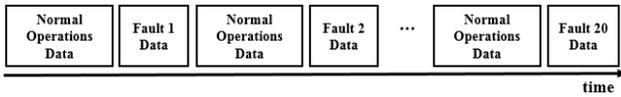

*Figure 1 – Phase II Data*

The fraction outlier *f* was set to 0.001 and a sample size of 42 was chosen for the sampling based method. Table 1 summarizes the performance of the $K_T$ charts. We were able to identify the presence of faults using either the $R^2$ or the *a* chart for all faults except fault 14, 15 and 19.

| Delay in Fault detection (# windows) | Fault Number |
|---|---|
| 1 | 4,5,7 |
| 2 | 1 |
| 3 | 6 |
| 4 | 16 |
| 5 | 2,3,10,11,12,13,18 |
| 6 | 8,9,14,15,19,20 |

*TABLE 1 – Performance Summary*

Figure 2 to 7 illustrate the $K_T$ chart results for faults 1, 4 and 19 respectively. These represent two results where $K_T$ charts could identify faults and one result where $K_T$ charts were not able to identify presence of faults.

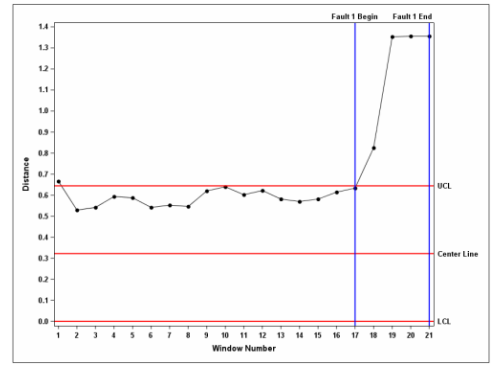

*Figure 2 – **a** chart for fault 1*

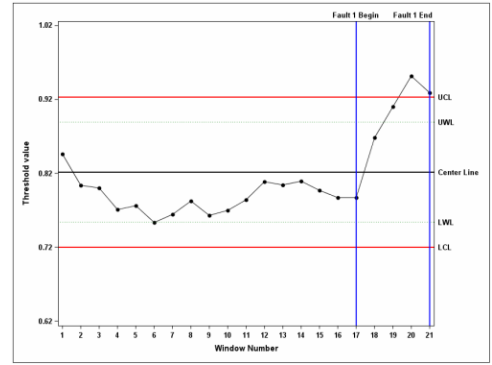

*Figure 3 - $R^2$ chart for fault 1*

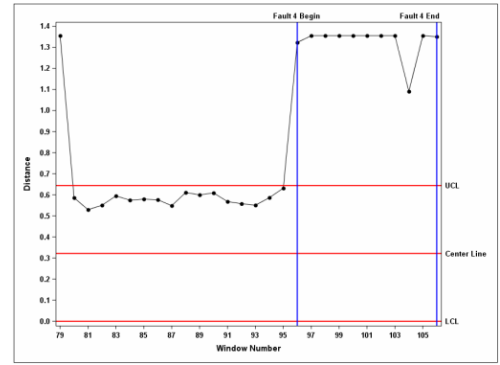

*Figure 4 – **a** chart for fault 4*

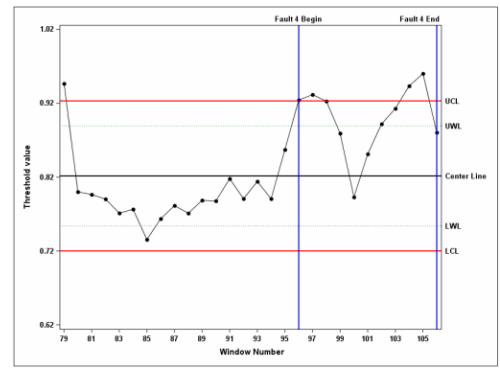

*Figure 5 - $R^2$ chart for fault 4*

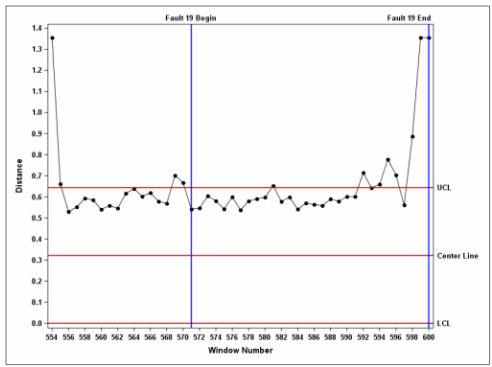

Figure 6 – *a* chart for fault 19

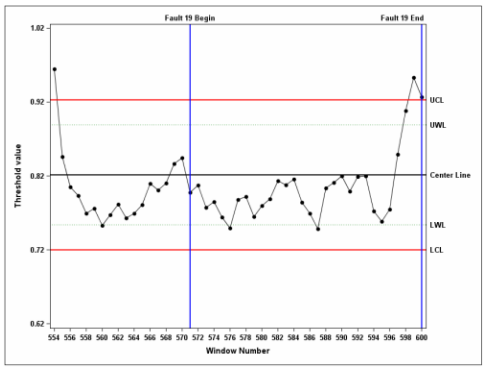

Figure 7 - $R^2$ chart for fault 19

## CONCLUSIONS

We proposed a new SVDD based $K_T$ control chart. The control chart can be used to simultaneously monitor the central tendency and the dispersion of multivariate data. The sampling method of SVDD training has a low memory requirement. This facilitates moving the $K_T$ control chart computations embedded in the physical asset being monitored or to the *edge*, as referred to in the IoT parlance. Since the multivariate sensor data generated from reliable machines, generally refers to a normal operating state, moving $K_T$ control chart computations to the edge can result in reduction in the data transmission costs in Phase II. This is possible by transmitting observations from the edge to the central server only when the corresponding window falls outside the control limits.

In future research we plan to develop a methodology to identify the variables responsible for the change in process center and spread. We are also actively researching a new process capability index, using the $K_T$ control chart statistics such as the threshold $R^2$ and the center *a*.